\documentclass[letterpaper]{article} % DO NOT CHANGE THIS
\usepackage{aaai25}  % DO NOT CHANGE THIS
\usepackage{times}  % DO NOT CHANGE THIS
\usepackage{helvet}  % DO NOT CHANGE THIS
\usepackage{courier}  % DO NOT CHANGE THIS
\usepackage[hyphens]{url}  % DO NOT CHANGE THIS
\usepackage{graphicx} % DO NOT CHANGE THIS
\usepackage{natbib}  % DO NOT CHANGE THIS AND DO NOT ADD ANY OPTIONS TO IT
\usepackage{caption} % DO NOT CHANGE THIS AND DO NOT ADD ANY OPTIONS TO IT
\usepackage{algorithm}
\usepackage{algorithmic}

\usepackage{newfloat}
\usepackage{listings}
\DeclareCaptionStyle{ruled}{labelfont=normalfont,labelsep=colon,strut=off} % DO NOT CHANGE THIS
\floatstyle{ruled}
\newfloat{listing}{tb}{lst}{}
\floatname{listing}{Listing}
\usepackage{amsmath}

\title{VidSole: A Multimodal Dataset for Joint Kinetics Quantification and Disease Detection with Deep Learning}
\author {
    Archit Kambhamettu\textsuperscript{\rm 1,}\footnote{co-first authors},
    Samantha Snyder\textsuperscript{\rm 1,}\footnotemark[1],
    Maliheh Fakhar\textsuperscript{\rm 1}, 
    Samuel Audia\textsuperscript{\rm 1},
    Ross Miller\textsuperscript{\rm 1},
    Jae Kun Shim\textsuperscript{\rm 1},
    Aniket Bera\textsuperscript{\rm 2}}
\affiliations {
    \textsuperscript{\rm 1}University of Maryland, College Park\\
    \textsuperscript{\rm 2}Purdue University\\

    \{architk, snyder36, mfakhar, sjaudia, rosshm, jkshim\}@umd.edu,  aniketbera@purdue.edu
}

\begin{document}

\maketitle

\begin{abstract}
Understanding internal joint loading is critical for diagnosing gait-related diseases such as knee osteoarthritis; however, current methods of measuring joint risk factors are time-consuming, expensive, and restricted to lab settings. In this paper, we enable the large-scale, cost-effective biomechanical analysis of joint loading via three key contributions: the development and deployment of novel instrumented insoles, the creation of a large multimodal biomechanics dataset (VidSole), and a baseline deep learning pipeline to predict internal joint loading factors. Our novel instrumented insole measures the tri-axial forces and moments across five high-pressure points under the foot. VidSole consists of the forces and moments measured by these insoles along with corresponding RGB video from two viewpoints, 3D body motion capture, and force plate data for over 2,600 trials of 52 diverse participants performing four fundamental activities of daily living (sit-to-stand, stand-to-sit, walking, and running). We feed the insole data and kinematic parameters extractable from video (i.e., pose, knee angle) into a deep learning pipeline consisting of an ensemble Gated Recurrent Unit (GRU) activity classifier followed by activity-specific Long Short Term Memory (LSTM) regression networks to estimate knee adduction moment (KAM), a biomechanical risk factor for knee osteoarthritis. The successful classification of activities at an accuracy of 99.02 percent and KAM estimation with mean absolute error (MAE) less than 0.5 percent*body weight*height, the current threshold for accurately detecting knee osteoarthritis with KAM, illustrates the usefulness of our dataset for future research and clinical settings.

\end{abstract}

% % Uncomment the following to link to your code, datasets, an extended version or similar.

\section{Introduction}

Over one in three Americans older than the age of 65 experience some form of physical disability \cite{Fuller-Thomson2023TemporalAmericans.}. These disabilities prevent them from performing activities of daily living (ADL), such as sitting, standing, walking, and running.  Osteoarthritis, the leading cause of disability in this age group, is characterized by joint structural changes and cartilage degeneration due to gait abnormalities that cause an uneven distribution of forces. \cite{Martel-Pelletier2016Osteoarthritis, Felson2013}. During an individual's lifetime, cyclical joint loading can result in osteoarthritis \cite{Miyazaki2002DynamicOsteoarthritis, Chehab2014BaselineOsteoarthritis}; however, altering gait to reduce joint loads can reduce symptoms of osteoarthritis \cite{Eddo2017CurrentReview, Shull2013Six-weekOsteoarthritis,Rynne2022EffectivenessMeta-analysis}. Therefore, the accurate measurement of the lower body kinetics (e.g., joint loads, moments) and kinematics (e.g., joint angles, speed) during ADL is vital in preventing joint-related diseases \cite{Sharma2010,Heiden2009,Fitzgerald2004,Winter2009,Labban2021KineticApproaches.}. 

Current methods to measure the kinetics and kinematics of joints require high-quality 3D motion capture systems and force plates, costing over \$150,000.  Furthermore, traditional gait analysis is time-consuming \cite{Hulleck2022PresentTechnologies.} and requires the guidance of trained biomechanists. While biomechanics experts have hinted at the potential transition from the tedious use of body markers and force plates to cost-effective wearables and video, they have emphasized the need for the proper development of accurate, lightweight models that can reduce the complexity of the problems for users of minimal technical expertise \cite{Stenum2024ClinicalChange}. Although this is an active area of research \cite{Molavian2023}, practical applications and streamlined approaches applicable to clinical settings do not exist yet. 

Recent methods to measure osteoarthritis-related risk factors propose AI applied to video datasets \cite{Uhlrich2023OpenCap:Videos} or wearable devices such as instrumented insoles \cite{Jacobs2015EstimationSensors, Savelberg1999a,Sim2015a}; however, these unimodal approaches are still time-consuming, require technical expertise, and need to be further validated in practical environments. 

We aim to foster developments in robust, easy-to-use approaches to estimating joint kinetics and kinematics by first designing a comparatively economical, tri-axial instrumented insole that measures the forces and moments across five pressure points on the foot. We then conduct a large-scale, multimodal gait analysis, consisting of 52 participants who each perform 25 walking, 10 walking with a sidestep, 10 sit-to-stand, 10 stand-to-sit, 15 running, and 6 standing trials on average. For each trial, we collect the corresponding force plate, 3D motion capture, instrumented insole, and 2-viewpoint RGB video data to create VidSole, our multimodal biomechanics dataset. 

\begin{figure*}[ht]
  \centering
  \includegraphics[scale = 0.80]{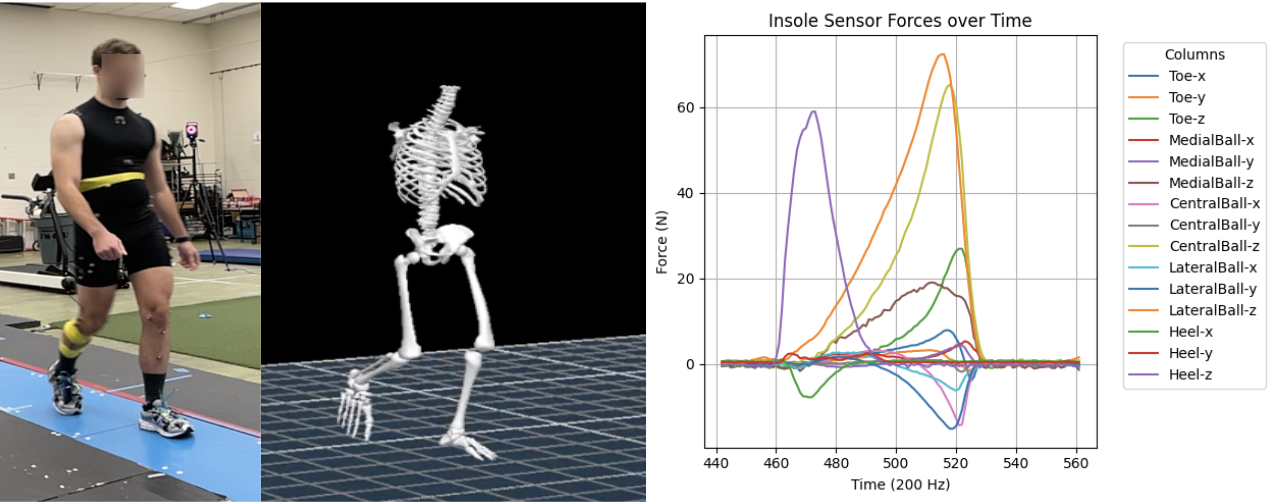}
   \caption{We assemble VidSole, a dataset that includes RGB video, motion capture, force plate, and instrumented insole pressure forces and moments data. This figure shows the RGB video data, motion capture visualized skeleton, and insole sensor raw force data for a participant walking.}
  \label{fig:teaser}
\end{figure*}

\begin{itemize}
  \item The design and deployment of a novel, reliable instrumented insole capable of measuring tri-axial forces and moments outside of a gait lab.
  \item Introduction of VidSole, a large multimodal dataset for biomechanical analysis, consisting of corresponding instrumented insole, 2-viewpoint RGB video, 3D motion capture, and force plate data for over 2,600 trials of 52 participants performing four fundamental activities of daily living (sit-to-stand, stand-to-sit, walking, and running). %This dataset is publicly available.%\footnote{\url{}}. 
  \item Development of a deep learning pipeline to accurately measure knee adduction moment (KAM), a biomechanical risk factor for knee osteoarthritis, and benchmarking evaluations that affirm our models' predictions are useful for clinical decision-making.

\end{itemize}

\section{Related Work}

\subsection{Video Analysis in Biomechanics}

Currently, methods that can accurately quantify joint kinematics from pose estimation coupled with attention-based architectures have been developed in research settings \cite{Cotton2022TransformingAnalysis}. Furthermore, streamlined methods to properly quantify walking stance cycles and time per cycle currently utilize ground truth body markers but can be easily replaced with high-quality video pose estimators \cite{Stenum2024ClinicalChange}. In contrast, current approaches to quantify joint kinetics from only video feeds \cite{Uhlrich2023OpenCap:Videos} or motion capture data \cite{Boswell2020} are time-consuming and are of low accuracy in comparison to our multimodal approach.

\subsection{Instrumented Insoles and Machine Learning in Biomechanics}
Instrumented insoles can increase the accessibility of gait analysis by providing information about the magnitude and location of pressure under the foot in a small, portable device. Combining deep learning with pressure sensing insoles, researchers have accurately predicted center of pressure trajectories \cite{Choi2018GaitSensors}, ground reaction forces \cite{Rouhani2010AmbulatoryDistribution, Savelberg1999a}, and a combination of the two \cite{Sim2015a}. However, there are limitations to current commercial insoles. The F-scan\footnote{\url{https://www.tekscan.com}} and Pedar\footnote{{\url{https://www.novelusa.com/pedar}}}  insoles measure pressures only in the vertical direction. Although vertical forces are the predominant force affecting joint dynamics, shear forces, the force parallel to the cross-section of the joint, %\textbf{Add in how shear forces are non vertical} 
also have a significant effect on joint kinetics \cite{Helseth2008}. Another component of joint dynamics is the free moment, typically measured by force plates, which provides localization of forces acting on the foot \cite{Milner2006}. 
Shear forces and free moments are necessary for gait lab calculation of dynamics, and including these measurements could result in improved predictions of joint dynamics. To address these limitations, we develop our instrumented insoles with tri-axial sensors that capture forces and moments in all three directions. 

\subsection{Multimodal Approaches in Biomechanics}
Limited research exists on combining video and instrumented insoles to predict gait kinetics and kinematics. Recent studies use strictly instrumented insole data to quantify gait kinematics and only use video data for ground truth labeling \cite{Chatzaki2021, Ngueleu2019}. Furthermore, when publicly available, the final dataset only contains instrumented insole data. In contrast, our dataset and pipeline incorporate both video and insole data. To the best of our knowledge, VidSole is the first publicly available dataset that incorporates four modalities useful for the development of deep learning methods for joint kinetics and kinematics approximations: instrumented insoles forces and moments, RGB video, 3D motion capture, and force plate data. 

\section{Dataset Curation}

\subsection{Insole Development}

We develop custom insoles for 8 shoe sizes (Women Size US 7 and 8, Men Size US 8, 9, 10, 11, 12, and 13), embedded with five 6-degree freedom piezoresistive sensors. We select Shokac Chip 6DoF-P100\footnote{\url{http://www.touchence.jp/en/products/chip04.html}}  due to their efficacy in previous research \cite{Snyder2023a}, lightweight nature (10.2 grams per sensor), and because they are the smallest 6DoF sensors (20x20x7.1mm) commercially available. 

These sensors can measure forces and moments in three dimensions and are calibrated within specified ranges. Forces in the anterior-posterior (X direction in Figure 2) and medial-lateral directions (Y direction in Figure 2) are calibrated between -20 and 20 Newtons, while forces in the vertical directions are calibrated between -100 and 100 Newtons. Moments across all three directions are calibrated between -350 and 350 Netwon-milimeters. Although calibrated within these ranges, these sensors can additionally measure without saturation outside those ranges.

\begin{figure}[ht]
    \centering
    \includegraphics[width=1\linewidth]{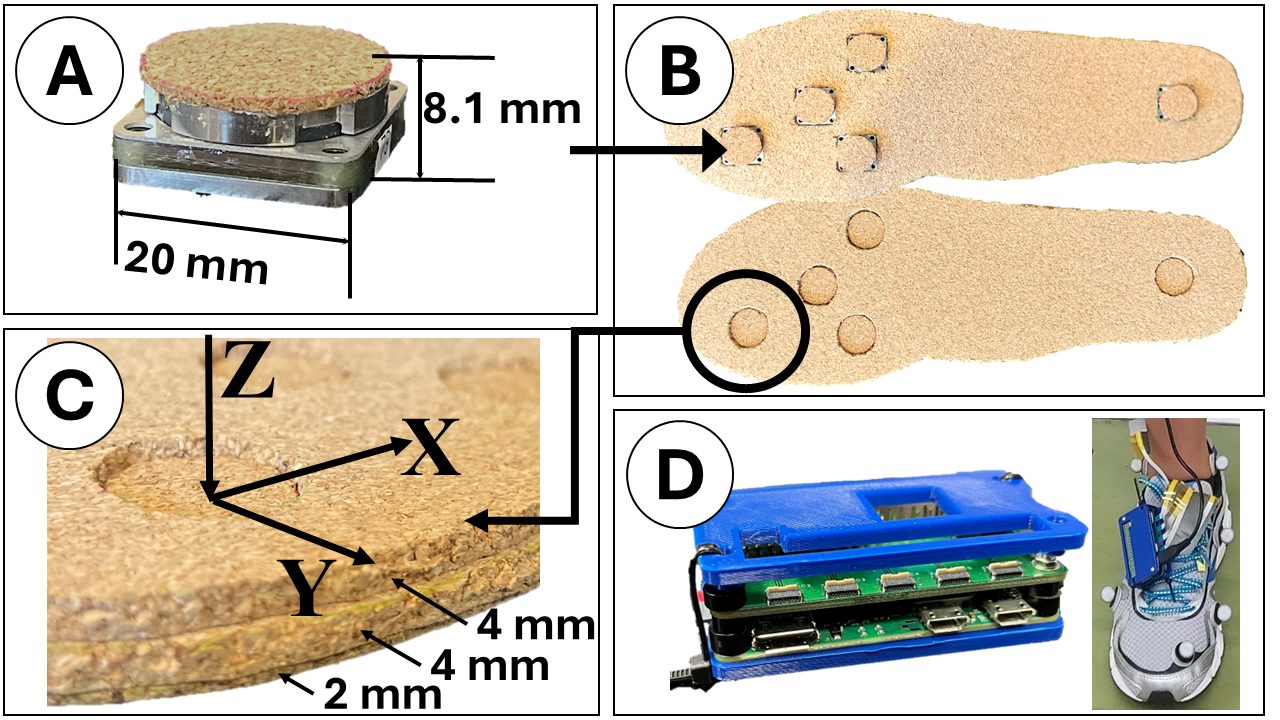}
    \caption{A) An individual sensor with a thin layer of cork increases its total height to 8mm. B) Sensors are aligned in the cork insole under the toe, medial ball, central ball, lateral ball, and heel. C) Axis of sensor orientation in the insole. The insole comprises two 4 mm pieces of cork and one 2 mm piece of cork. D) Insole in the shoe with Raspberry Pi and custom PCB housed on the top of the shoe.}
    \label{fig:insole_dev}
\end{figure}

We placed our sensors under five high-pressure areas of the foot: big toe, medial ball, lateral ball, central ball, and heel (Figure \ref{fig:insole_dev}). The location of the sensors was linearly scaled across shoe sizes to ensure they were in the same relative position for each shoe, and each sensor was embedded in the same axis-aligned orientation within the insole. We constructed the 10 mm thick right-foot insoles out of the cork and embed the sensors 8 mm into the cork to minimize stress. Each sensor was covered with 1 mm of cork to prevent user discomfort. We built a similar cork for the left foot without sensors to limit potential non-physiological asymmetries during movement.

We connected our sensors to a custom printed circuit board (PCB), attached to a Raspberry Pi Zero W via flat flex cables and collected data locally at 82 Hz with a script developed with an open source library pigpio\footnote{\url{https://abyz.me.uk/rpi/pigpio/}}. The total combined weight of the Raspberry Pi Zero W and PCB was negligible (28 grams) and was secured to the top of the shoe \ref{fig:insole_dev}. We connected the Raspberry Pi Zero W via a battery pack strapped to the waist of each participant, with the cable secured to each participant's leg. We synchronized the vertical forces post hoc with a cross-correlation method \cite{Spencer2023,Grouvel2023}.

During most daily activities, particularly running, the sensors were often exposed to values outside their working range. We confirmed that the sensors measured these values with moderate to excellent reliability during standing and running. Additionally, we established criterion validity, as there was an overall moderate correlation between the ground reaction force data and the insole sensor measurements during running.

\begin{figure*}[ht]
  \begin{center}
    \includegraphics[scale = 0.35]{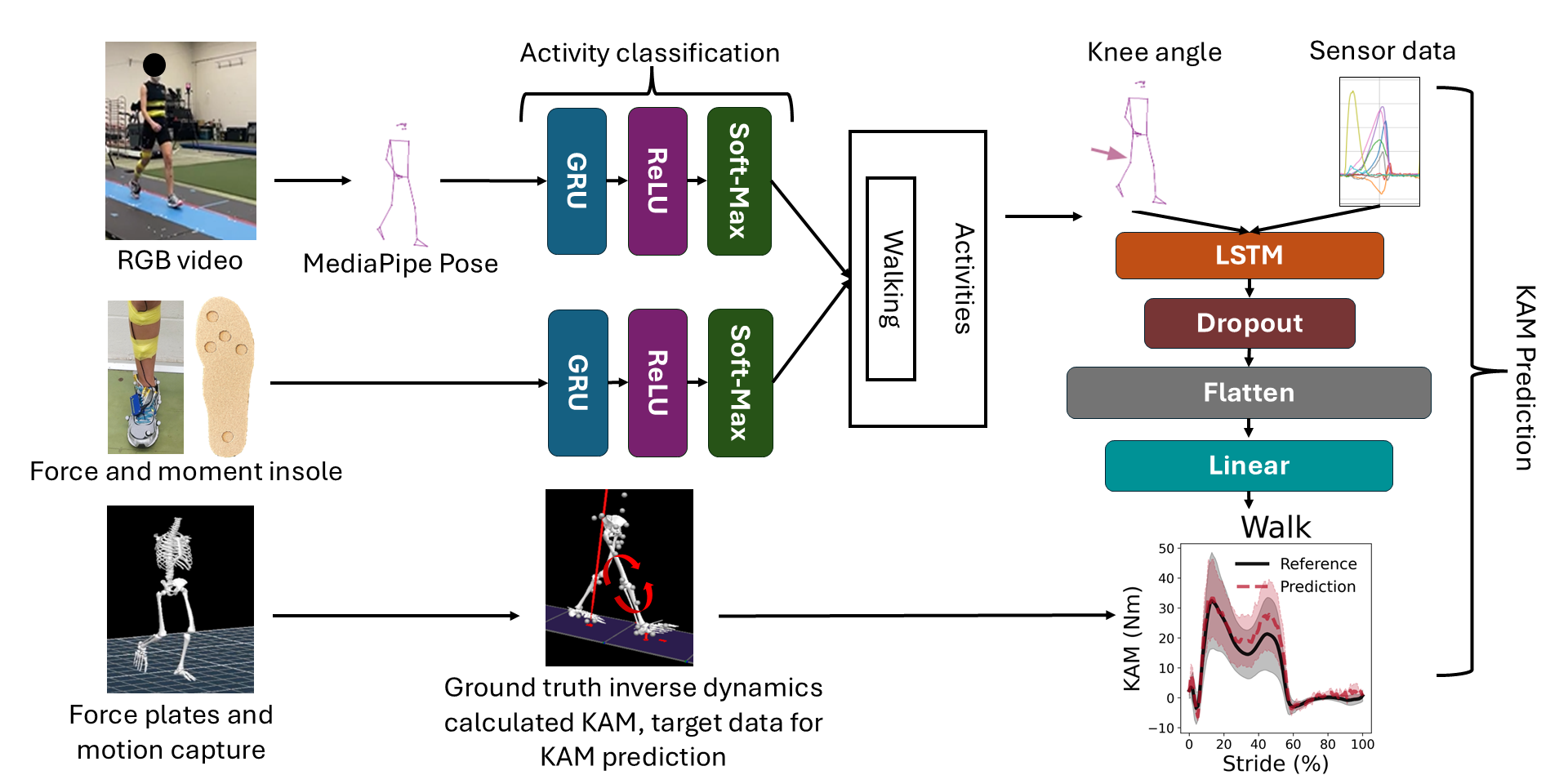}
  \end{center}
     \caption{Our deep learning pipeline: The insole data and MediaPipe pose estimation extracted from RGB video are used to classify activities using an ensemble Gated Recurrent Unit (GRU) model. After each activity is classified, the insole sensor data and knee angle data are used as inputs to an activity-specific Long Short Term Memory (LSTM) model to predict KAM. Ground truth KAM is calculated via inverse dynamics from the force plate and motion capture data.}
  \label{fig:workflow}
\end{figure*}

\subsection{Data Collection}
52 healthy, young male and female individuals provided informed consent of protocol (Institutional Review Board \#1335286). Participants then changed into tight-fitting clothing and custom instrumented shoes fitted to the nearest whole shoe size. We collect gait data with thirteen motion capture cameras (200 Hz; T-Series; Vicon Inc.\footnote{\url{https://www.vicon.com/}}, Oxford, UK) and seven force plates along a 12-meter walkway (1000 Hz; 9260AA, Kistler Instrument Corp\footnote{\url{https://www.kistler.com/US/en/}}.). 
First, participants performed two variations of static trials: one with their arms in a T-pose, and one with their arms crossed. The first variation was performed once as a calibration trial, while the second was performed 3 times and recorded. In both, participants stepped onto two force plates, one foot positioned shoulder width apart on both plates, and stood still with arms in the designated orientation for 30 seconds. Participants then performed 3 trials with their arms crossed and their left foot (foot without insole) off the ground at a 90-degree angle for 10 seconds. Following these, participants performed sit-to-stand and stand-to-sit trials, where they were instructed to sit and stand from a stool, arms crossed, at a comfortable pace 10 times, with a short rest between each trial. For the walking trials, participants were instructed to walk across a 12-meter force plate walkway at a comfortable pace for 10 trials, walk at a fast pace for 5 trials, walk at a slow pace for 5 trials, and walk at a pace of 1.3 m/s for 5 trials. Following this, participants performed 10 side-step trials where they walked at a self-selected speed, planted their right foot, and stepped to the left. Lastly, participants were instructed to run at a comfortable self-selected pace for 5 trials, run at a pace of 3 m/s for 5 trials, and run at various speeds for 5 trials. The order of speeds for walking and running was randomized for each participant. 

We leverage the Vicon Nexus\footnote{\url{https://www.vicon.com/software/nexus/}} software to label our 3D motion capture marker data and import this labeled data into Visual3D\footnote{\url{https://www.has-motion.ca/}} for ground truth KAM calculations. We remove high-frequency noise from our marker data by filtering our running marker data at 10 Hz and other activity data at 6 Hz via a dual pass 4th-order Butterworth filter \cite{Hunter2019FastRunning, Krupenevich2015MalesLoad}. We filter ground reaction force with a cutoff frequency of 45 Hz via a dual pass 4th-order Butterworth filter as well \cite{Krupenevich2015MalesLoad}. Based on the calibration trial, a model of the participant is created and KAM is calculated via inverse dynamics in Visual3D relative to the tibia. 

\begin{table}[ht]

\begin{center}
\begin{tabular}{|c|c|}
\hline
\multicolumn{2}{|c|}{\bfseries Participants (N=52)} \\
\hline
\hline
\bfseries  Gender & 26 Male, 26 Female \\
\hline
\bfseries  Age (years) & 23.4$\pm$4.0 \\
\hline
\bfseries  Height (cm) & 168.9$\pm$8.7 \\
\hline
\bfseries  Weight (kg) & 65.6$\pm$9.9 \\
\hline
\bfseries   & 24 White \\
\bfseries   & 22 Asian \\
\bfseries  Race & 3 African American\\
\bfseries   & 3 Hispanic/Latinx \\
\bfseries   & 3 Middle Eastern \\
\hline
\end{tabular}
\end{center}
\caption{Participant demographics.}
\label{tab:participant_demographics1}
\end{table}

\begin{table}[ht]

\begin{center}
\begin{tabular}{|c||c |c |c|}
\hline
& \bfseries  Insole  &   \bfseries Motion & \bfseries  RGB Video \\
&  & \bfseries  Capture + &   \\
&  & \bfseries  Force Plate &   \\
\hline
\hline
& \multicolumn{3}{|c|}{ \bfseries Standing (2-Leg and 1-Leg combined)} \\
\hline
\bfseries Subjects & 47 &  52 & 41 \\
\hline
\bfseries Trials & 6 &  6 & 6 \\
\hline
\bfseries Frames & 462,480 & 1,248,000 & 147,600\\
\hline
& \multicolumn{3}{|c|}{\bfseries Sit-to-Stand and Stand-to-Sit (each)} \\
\hline
\bfseries Subjects & 47 &  52 & 41 \\
\hline
\bfseries Trials & 10 &  10 & 10 \\
\hline
\bfseries Frames & 231,240 & 624,000 & 615,000 \\
\hline
& \multicolumn{3}{|c|}{\bfseries Walking} \\
\hline
\bfseries Subjects  & 47 &  52 & 41 \\
\hline
\bfseries Trials & 25 &  25 & 25 \\
\hline
\bfseries Frames & 481,750 & 1,300,000 & 307,500 \\
\hline
& \multicolumn{3}{|c|}{\bfseries Walking with Side-Step} \\
\hline
\bfseries Subjects  & 47 &  52 & 41 \\
\hline
\bfseries Trials & 10 &  10 & 10 \\
\hline
\bfseries Frames & 192,700 & 520,000 & 123,000 \\
\hline
& \multicolumn{3}{|c|}{\bfseries Running} \\
\hline
\bfseries Subjects  & 47 &  52 & 0 \\
\hline
\bfseries Trials & 15 &  15 & 0 \\
\hline
\bfseries Frames & 173,430 & 468,000 & 0\\
\hline
\end{tabular}
\end{center}
\caption{Data Breakdown}
\label{tab:data_descriptions}
\end{table}
\subsection{Data Statistics}
Our VidSole dataset consists of 52 data collections, and participant demographics are presented in Table \ref{tab:participant_demographics1}, with multiracial participants listed under each of their races. 
The dataset consists of 76 trials per participant on average and 2,632 valid trials in total. The data breakdown is described in Table \ref{tab:data_descriptions}, and specific trial information is detailed in the Data Collection section. During each session, sensor and video data were recorded for all trials except running, where video data was not recorded. Standing still trials are 30 seconds and standing on one leg trials are 10 seconds. Sit-to-stand and stand-to-sit trials were recorded together and were approximately 6 seconds in length. Walking trials occur at various speeds but are approximately 5 seconds in length on average, and similarly, running trials are approximately 3 seconds in length on average. There are fewer usable trials for the joint kinetic prediction in comparison to activity classification because force plate data is unusable for the ground truth kinetic calculations when two feet are on one force plate. Data was collected with the insoles at 82 Hz, the synchronized motion capture and force plate system at 200 Hz, and RGB video data at 60 frames per second (FPS). Sensor data is provided as binary files, motion capture and force plate data is a c3d file type, and RGB video file is an mp4 file type. This data and corresponding code are accessible via GitHub: (to be included upon acceptance). 

\section{Deep Learning for Knee Joint Kinetics}
We provide a baseline approach for predicting knee joint kinetics, specifically the knee adduction moment (KAM), for each activity of daily living (ADL). Providing automated preprocessing \cite{Ozates2024IdentificationAI}, reducing model complexity, and minimizing reliance on large computational resources \cite{Stenum2024ClinicalChange} allows users of minimal technical experience to utilize these systems. Thus, we devise a fully automated pipeline that requires minimal user-based preprocessing to provide accurate joint kinetic metrics. We do so by dividing the task into two sub-portions: Activity Classification and Kinetics Prediction.

\subsection{Activity Classification}

We introduce this activity classification model as a temporal grounding step to effectively extract each important class (sit-to-stand, stand-to-sit, walking, running) from the other labeled and unlabeled (eg swaying, strictly standing/sitting) actions. Furthermore, activity classification allows for the tuning of activity-specific KAM estimation models. The ensemble model consists of two Gated Recurrent Units (GRU) models, one for the video input stream and one for the insole sensor input stream. We select GRU \cite{chung2014empiricalevaluationgatedrecurrent} as our final model after evaluating the efficacy of the LSTM \cite{Sherstinsky_2020} and vanilla Recurrent Neural Network \cite{lipton2015criticalreviewrecurrentneural} models for this task. We demonstrate high success via the GRU architecture (hidden layer size 16), so we do not consider heavier architectures like the Transformer \cite{vaswani2023attentionneed}, to enable our models to process quickly in real-world scenarios with minimal compute resources. Furthermore, we take an ensemble approach to this problem to be robust in real-world scenarios where data may be corrupted; our ensemble model automatically makes predictions via the available input streams. 
We use a sequence length of 60 for the video model and 82 for the insole model, leveraging a sliding window approach with an overlap length of 50 and 70, respectively. The 52 participants were divided into a 70\% / 13\% / 17\% training / validation/ testing split, to ensure that the data of the participants were not included in more than one set \cite{Halilaj2018MachineOpportunities}. Due to the imbalanced nature of the dataset, we utilize a weighted random sampler during training to increase the probability of sampling instances from the minority classes (sit-to-stand and stand-to-sit) for each batch, thereby ensuring a better representation of the sample space during training. To further counteract the imbalanced nature of the dataset, we leverage the Weighted Cross Entropy Loss \cite{Ozdemir2020WeightedImages}. For both models, we utilize the adaptive momentum optimizer (ADAM) \cite{kingma2014adam} with a learning rate of 0.003 and train with 1 RTX A4000 GPU. 

\subsubsection{Classification Ensemble Model}
The first classification sub-model receives RGB video data as input, detailed in Figure \ref{fig:workflow}. We take a pose-based activity classification approach, which is more robust to changes in the background during inference \cite{Singh2023FastClassifiers}, to ensure our approach is generalizable beyond the scope of our dataset \cite{Thilakarathne2022}. We use the out-of-the-box pose estimation model Mediapipe \cite{Lugaresi2019MediaPipe:Pipelines} due to its ease of use for users with minimal technical experience, limited computational resource requirements, and its use in the biomechanics literature \cite{Pattanapisont2024Multi-ViewParts}. For each frame in the video, we extract 14 keypoints out of 33 potential keypoints (left and right wrists, elbows, shoulders, hips, knees, ankles, and index toes) from pose estimation and concatenate them across all frames. Figure \ref{fig:workflow} details the video classification model, which we train for 80 epochs, employing early-stopping (patience = 5) to prevent overfitting. 

To limit the need for user preprocessing, we feed raw sensor data streams directly from our instrumented insoles to our second classification model, detailed in Figure \ref{fig:workflow}. We train this model with a batch size of 64 for 100 epochs with early-stopping (patience = 5).

\subsubsection{Activity Classification Performance}
The video and insole models are trained independently, and their resulting probability scores are averaged during inference. The ensemble approach results in an accuracy of $99.02\%$ on our test set. We perform an ablation study by comparing the two independent single-mode models. Although both the insole-only model ($97.74\%$ accuracy) and pose-based video model ($98.28\%$ accuracy) perform well, neither performs as well as the ensemble approach. 

\begin{table}[ht]
\begin{center}
\begin{tabular}{|c||c|c|c|}
\hline
\multicolumn{4}{|c|}{\textbf{Sensor Data Input}} \\
\hline
& $\pmb{r}$ & \multicolumn{2}{|c|}{\textbf{MAE}} \\
\hline
& & \textbf{Nm} & \textbf{\%BW*ht} \\
\hline
\hline
\textbf{Walking} & 0.94$\pm$0.05  & 4.37$\pm$1.93  & 0.42$\pm$0.18 \\
\hline
\textbf{Running} & 0.92$\pm$0.16 & 5.63$\pm$2.91 & 0.55$\pm$0.28 \\
\hline
\textbf{Stand-to-Sit} & 0.40$\pm$0.52 & 4.58$\pm$3.49 & 0.44$\pm$0.33 \\
\hline
\textbf{Sit-to-Stand} & 0.54$\pm$0.42 & 3.69$\pm$2.03 & 0.35$\pm$0.19 \\
\hline
\cline{1-4}
\multicolumn{4}{|c|}{\textbf{Sensor and Knee Angle Data Input}} \\
\cline{1-4}
& $\pmb{r}$ & \multicolumn{2}{|c|}{\textbf{MAE}} \\
\hline
& & \textbf{Nm} & \textbf{\%BW*ht} \\
\hline
\hline
\textbf{Walking} & 0.94$\pm$0.04 & 3.92$\pm$1.65 & 0.37$\pm$0.16 \\
\hline
\textbf{Running} & 0.94$\pm$0.13 & 4.72$\pm$1.68 & 0.46$\pm$0.16 \\
\hline
\textbf{Stand-to-Sit} & 0.64$\pm$0.28 & 3.27$\pm$2.25 & 0.31$\pm$0.21 \\
\hline
\textbf{Sit-to-Stand} & 0.64$\pm$0.40 & 3.20$\pm$1.78 & 0.30$\pm$0.17 \\
\hline
\end{tabular}
\end{center}

\caption{Kinetic model KAM prediction correlation coefficients ($r$) and mean average error (MAE) for each activity. MAE presented in Newton meters (NM) and normalized to bodyweight and height (\%BW*ht). Predictions are shown using only sensor data as input to a kinetic model and using sensor and pose-estimation calculated knee angles as inputs to the kinetic model. Results presented as mean $\pm$ one standard deviation.}
\label{table_kinetic_kneeAng1}
\end{table}

\begin{figure}[ht]
  \centering
  \includegraphics[width=1\linewidth]{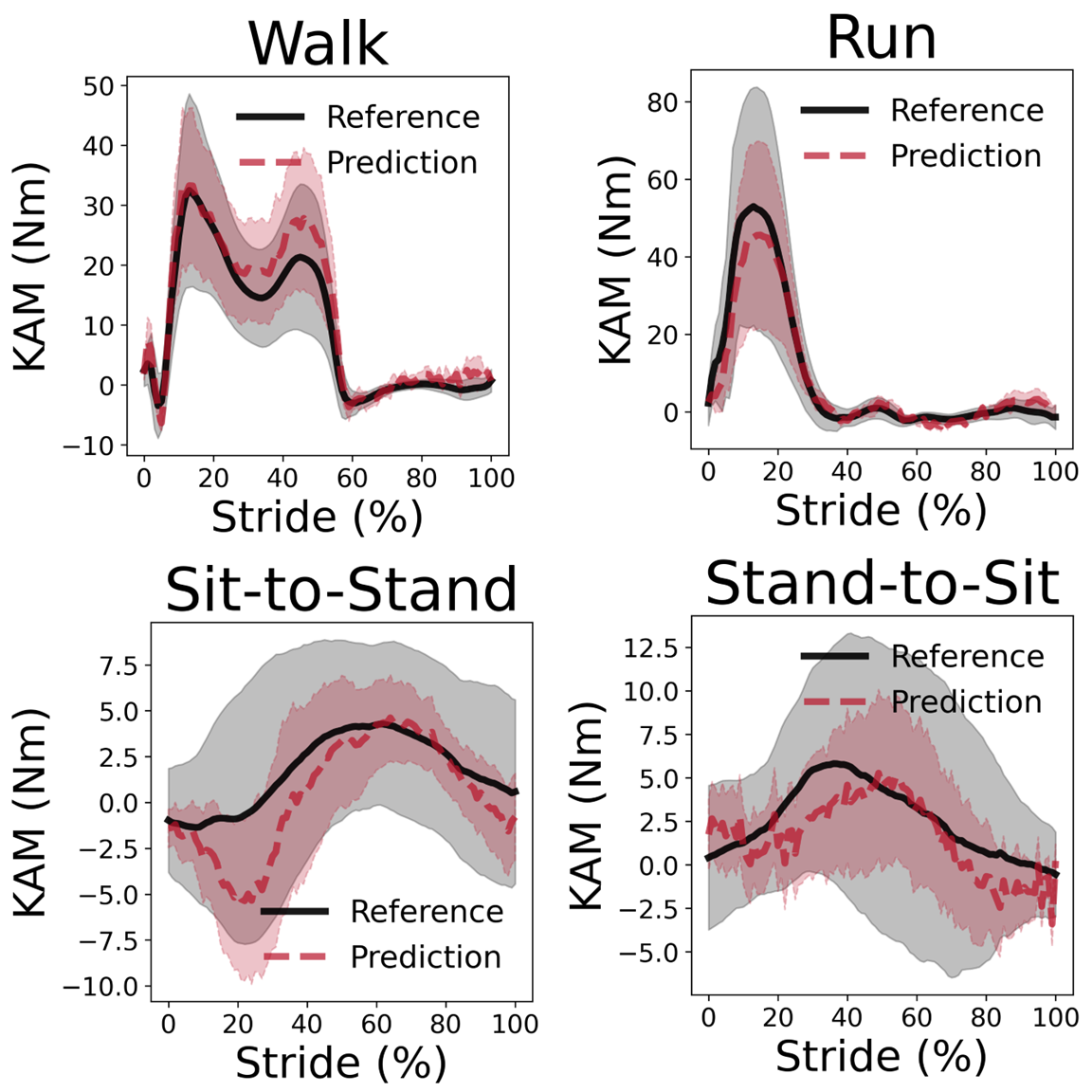}
  \caption{Visualization of mean KAM prediction (red dashed line) and mean reference ground truth (black line) for each activity model for multimodal inputs. One standard deviation is plotted in the shaded area.}
  \label{fig:KAMplots}
\end{figure}

\subsection{Knee Joint Kinetics}

After classifying activities, dynamic measures, such as KAM, can be predicted using regression models with sensor and video data as input.  The 52 participants are split into a 70\%, 13\%, 17\%, train, validation, test split, and the participants' data is not included in more than one of the sets \cite{Halilaj2018MachineOpportunities}. We extracted each stride and corresponding knee angles from motion data. Each data stream is scaled to 101-time points to represent 0 to 100\% of the stride, a typical way to examine data in biomechanics \cite{Winter2009}.

After preliminary tests with baseline versions of four commonly used models when predicting KAM, multilayer perceptron, recurrent neural networks (RNN), and convolutional neural networks \cite{Boswell2020,Snyder2023a}, we found that recurrent models performed the best. Therefore, we first develop LSTM models \cite{Sherstinsky_2020} with forces from the 6 degrees of freedom insole data as input to estimate KAM during walking, running, sit-to-stand, and stand-to-sit activities. We then improve on these models and demonstrate the efficacy of multimodal approaches by developing models that predict KAM from both insole data and the time-aligned, pose-derived knee angle data.

\subsubsection{Model Description}
We first develop LSTM models for each class (walking, running, sit-to-stand, and stand-to-sit) with only force-insole sensor data as input. Raw sensor data serves as the input to the LSTM, where each input consists of 15 sensor features (forces in X, Y, and Z directions) and 101 time points. The predicted KAM is then compared to ground truth, gait lab calculated KAM, and the basic LSTM is fine-tuned using Ray Tuner\footnote{\url{https://docs.ray.io/en/latest/tune/index.html}} to minimize mean-average-error loss of KAM across strides. Each LSTM consists of an LSTM layer, followed by a fully connected ReLu layer, and a fully connected linear layer as described in Figure \ref{fig:workflow}. We use a batch size of 10 for running and sitting, and a batch size of 20 for walking. The LSTM layer size differs by activity, with a size of 128 for walking and running and a size of 256 for sitting and standing. The models leverage the ADAM optimizer \cite{kingma2014adam} with a learning rate of 0.08 and are trained for 150 to 300 epochs on a single NVIDIA GeForce RTX 4070 Ti GPU. We utilize a scheduler to reduce the learning rate on a plateau and include dropout layers in the model to prevent overfitting. 

To demonstrate the ability of video data to increase the prediction accuracy of kinetic parameters, specifically KAM, we create a second baseline LSTM model for each activity class. The secondary model intakes the pose-calculated knee angle from the time-aligned RGB video data coupled with the insole sensor data as input. Each secondary model has the same layers as the original model, with a fine-tuned number of nodes and learning rate. 
 
 \subsubsection{Kinetic Model Performance}
As demonstrated in Table \ref{table_kinetic_kneeAng1}, our multi-modal model predicts KAM with mean average errors (MAE) lower than the 0.5 \%BW*ht threshold, thus demonstrating its potential use in clinical settings as a viable estimator of KAM to detect early signs of knee osteoarthritis. In comparing correlation coefficients, the walking and running models significantly outperform the sit-to-stand and stand-to-sit models, both with and without the addition of the knee angle. This could be due to the greater noise-to-signal ratio and less variation in sensor magnitudes across the sitting/standing tasks. Additionally, the sit-to-stand and stand-to-sit results demonstrate the importance of a multimodal dataset, as including the knee angle greatly increased the correlation coefficients of the model. 

Our walking model ($r=0.94$) significantly outperforms the best video-only (OpenCap $r^2=0.80$) \cite{Uhlrich2023OpenCap:Videos}, and motion-capture-only ($r^2 = 0.86$) \cite{Boswell2020} approaches to KAM estimation, further emphasizing the robustness of our multimodal approach. Overall, our baseline models show promising results, and future research incorporating additional kinematic parameters from video data (joint speeds, hip angles, etc.) has the potential to further increase prediction accuracies.

\section{Conclusion}
In this paper, we propose VidSole, a novel dataset that includes 52 diverse individuals completing over 2,600 walking, running, sitting, and standing trials consisting of RGB video, instrumented insole force and moment, motion capture, and force plate data. This dataset, coupled with deep learning architectures, can facilitate advanced gait analysis in clinical settings. We illustrate this by introducing a pipeline to compute KAM, a biomechanical risk factor for knee osteoarthritis. In our pipeline, we first classify activities with high accuracy using a pose estimation and insole data-based GRU ensemble model. Our pipeline then predicts KAM via activity-specific LSTM regression models with only insole data and knee angles as inputs. Our LSTM regression models estimate KAM with moderate to high correlation coefficients, $0.64<r<0.94$, for all models, outperforming current methods and emphasizing the efficacy of our approach. Furthermore, our MAE is lower than the threshold for clinically applicable estimations, 0.5\%BW*ht \cite{Mundermann2004,Amin2004,Sharma1998KneeOsteoarthritis,Miyazaki2002DynamicOsteoarthritis}, further validating our approach. 
\subsubsection{Dataset Directions}
Given our success in developing models to estimate KAM, we strongly believe VidSole can be leveraged by the AI community to accurately estimate kinetic measures related to knee osteoarthritis and significant gait parameters related to other pathologies. Thus, we introduce two prevalent biomechanical measures that can be explored with our multimodal dataset. 

Shear forces (anterior-posterior forces parallel to the ground) acting on the knee are associated with the magnitude of ACL loading \cite{Maniar2022MuscleLoading}, ACL injury \cite{Yu2007}, and cartilage deterioration \cite{Lynn2007TheAnalysis}. The current means of calculating these forces are limited to utilizing inverse dynamics methods on motion capture and force plate data \cite{Winter2009}. As our novel instrumented insole measures forces and moments in non-vertical directions, we believe that coupled with proper spacial localization from RGB video, deep learning methodologies can produce accurate estimations for shear forces and aid in the early detection of knee-related diseases. 

The sit-to-stand test is one of the most popular methods for evaluating lower muscular power. The time required to complete 10 sit-to-stand repetitions is associated with lower extremity muscle strength; the number of sit-to-stand repetitions completed within 30 seconds relates to lower extremity muscle endurance; and the time to complete 5 sit-to-stand repetitions is associated with knee extension strength in adults of all age groups \cite{Lein2022Normative, Yoshiko2021, Bohannon2010Sit-to-standAge-span.,Atrsaei2022InstrumentedTest.}. The instrumented sit-to-stand test, in which each cycle of the sit-to-stand process is segmented and time per period is calculated, has a far higher association with a health status and functional ability \cite{vanLummel2016TheAdults.}. Current methods to measure this are via inertial measurement unit sensors placed near the center of mass; however, we believe our VidSole dataset can facilitate a more accurate, non-intrusive approach. Furthermore, 
 our multimodal dataset provides the necessary input for models to predict relevant biomechanical factors such as range of motion, angular velocity, coordination, and balance across different subphases of the test to further analyze the already proven sit-to-stand test. 
\subsubsection{Limitations and Future Work}
Although we attempt to mimic clinical settings as best as we can in our gait lab, further evaluating our methods in clinical settings is necessary for proper validation. While we detail the reasoning behind the technical decisions we made (video/insole alignment, deep learning pipeline, etc), we believe the development of more robust models that still adhere to the principles outlined in the paper (minimal preprocessing, automated pipeline, minimal computational requirement, etc.) is necessary for stronger results. Hence, we strongly believe the AI community will benefit from our dataset and baseline pipeline to optimize our approaches further. 

In our work, we explore how our dataset can be utilized to create large-scale gait analysis in clinical settings; however, the ability to track these important kinetic and kinematic metrics over the course of a day in real-world scenarios is still largely unexplored. Thus, we aim to test our instrumented insoles in all types of terrain and in non-supervised environments. We also plan to further diversify our dataset by collecting data on older individuals and individuals with knee osteoarthritis. This could lead to an advancement in medical decision-making and provide care for people of all ages, and assist in in fall detection, joint deterioration projections, and overall health quality. 

Thus, we emphasize the need for biomechanics, clinicians, and computer science researchers to work together to devise practical approaches to quantifying joint kinetics and preventing biomechanical-related diseases.

\bibliography{references, references1}

\begin{thebibliography}{56}
\providecommand{\natexlab}[1]{#1}

\bibitem[{Amin et~al.(2004)Amin, Luepongsak, McGibbon, LaValley, Krebs, and Felson}]{Amin2004}
Amin, S.; Luepongsak, N.; McGibbon, C.~A.; LaValley, M.~P.; Krebs, D.~E.; and Felson, D.~T. 2004.
\newblock {Knee adduction moment and development of chronic knee pain in elders}.
\newblock \emph{Arthritis Care and Research}, 51(3): 371--376.

\bibitem[{Atrsaei et~al.(2022)Atrsaei, Paraschiv-Ionescu, Krief, Henchoz, Santos-Eggimann, B{\"{u}}la, and Aminian}]{Atrsaei2022InstrumentedTest.}
Atrsaei, A.; Paraschiv-Ionescu, A.; Krief, H.; Henchoz, Y.; Santos-Eggimann, B.; B{\"{u}}la, C.; and Aminian, K. 2022.
\newblock {Instrumented 5-Time Sit-To-Stand Test: Parameters Predicting Serious Falls beyond the Duration of the Test.}
\newblock \emph{Gerontology}, 68(5): 587--600.

\bibitem[{Bohannon et~al.(2010)Bohannon, Bubela, Magasi, Wang, and Gershon}]{Bohannon2010Sit-to-standAge-span.}
Bohannon, R.~W.; Bubela, D.~J.; Magasi, S.~R.; Wang, Y.-C.; and Gershon, R.~C. 2010.
\newblock {Sit-to-stand test: Performance and determinants across the age-span.}
\newblock \emph{Isokinetics and exercise science}, 18(4): 235--240.

\bibitem[{Boswell et~al.(2020)Boswell, Uhlrich, Kidzinski, Thomas, Kolesar, Gold, Beaupre, and Delp}]{Boswell2020}
Boswell, M.~A.; Uhlrich, S.~D.; Kidzinski, L.; Thomas, K.; Kolesar, J.~A.; Gold, G.~E.; Beaupre, G.~S.; and Delp, S.~L. 2020.
\newblock {A neural network to predict the knee adduction moment in patients with osteoarthritis using anatomical landmarks obtainable from 2D video analysis}.
\newblock \emph{Osteoarthritis and Cartilage}, 29(3): 346--356.

\bibitem[{Chatzaki et~al.(2021)Chatzaki, Skaramagkas, Tachos, Christodoulakis, Maniadi, Kefalopoulou, Fotiadis, and Tsiknakis}]{Chatzaki2021}
Chatzaki, C.; Skaramagkas, V.; Tachos, N.; Christodoulakis, G.; Maniadi, E.; Kefalopoulou, Z.; Fotiadis, D.~I.; and Tsiknakis, M. 2021.
\newblock {The Smart-Insole Dataset: Gait Analysis Using Wearable Sensors with a Focus on Elderly and Parkinson’s Patients}.
\newblock \emph{Sensors}, 21(8).

\bibitem[{Chehab et~al.(2014)Chehab, Favre, Erhart-Hledik, and Andriacchi}]{Chehab2014BaselineOsteoarthritis}
Chehab, E.~F.; Favre, J.; Erhart-Hledik, J.~C.; and Andriacchi, T.~P. 2014.
\newblock {Baseline knee adduction and flexion moments during walking are both associated with 5year cartilage changes in patients with medial knee osteoarthritis}.
\newblock \emph{Osteoarthritis and Cartilage}, 22(11): 1833--1839.

\bibitem[{Choi et~al.(2018)Choi, Lee, Park, and Kim}]{Choi2018GaitSensors}
Choi, S.; Lee, S.; Park, H.; and Kim, H. 2018.
\newblock {Gait Type Classification Using Smart Insole Sensors}.
\newblock In \emph{TENCON 2018 - 2018 IEEE Region 10 Conference}, 1903--1906.
\newblock ISBN 2159-3450 VO -.

\bibitem[{Chung et~al.(2014)Chung, Gulcehre, Cho, and Bengio}]{chung2014empiricalevaluationgatedrecurrent}
Chung, J.; Gulcehre, C.; Cho, K.; and Bengio, Y. 2014.
\newblock Empirical Evaluation of Gated Recurrent Neural Networks on Sequence Modeling.
\newblock arXiv:1412.3555.

\bibitem[{Cotton et~al.(2022)Cotton, Mcclerklin, Cimorelli, Patel, and Karakostas}]{Cotton2022TransformingAnalysis}
Cotton, R.~J.; Mcclerklin, E.; Cimorelli, A.; Patel, A.; and Karakostas, T. 2022.
\newblock {Transforming Gait: Video-Based Spatiotemporal Gait Analysis}.
\newblock \emph{Proceedings of the Annual International Conference of the IEEE Engineering in Medicine and Biology Society, EMBS}, 2022-July: 115--120.

\bibitem[{Eddo et~al.(2017)Eddo, Lindsey, Caswell, and Cortes}]{Eddo2017CurrentReview}
Eddo, O.; Lindsey, B.; Caswell, S.~V.; and Cortes, N. 2017.
\newblock {Current Evidence of Gait Modification with Real-time Biofeedback to Alter Kinetic, Temporospatial, and Function-Related Outcomes: A Review}.
\newblock \emph{International Journal of Kinesiology and Sports Science}, 5(3): 35.

\bibitem[{Felson(2013)}]{Felson2013}
Felson, D.~T. 2013.
\newblock {Osteoarthritis as a disease of mechanics}.
\newblock \emph{Osteoarthritis and Cartilage}, 21(1): 10--15.

\bibitem[{Fitzgerald, Piva, and Irrgang(2004)}]{Fitzgerald2004}
Fitzgerald, G.~K.; Piva, S.~R.; and Irrgang, J.~J. 2004.
\newblock {Reports of Joint Instability in Knee Osteoarthritis: Its Prevalence and Relationship to Physical Function}.
\newblock \emph{Arthritis {\&} Rheumatism}, 51(6): 941--946.

\bibitem[{Fuller-Thomson, Ferreirinha, and Ahlin(2023)}]{Fuller-Thomson2023TemporalAmericans.}
Fuller-Thomson, E.; Ferreirinha, J.; and Ahlin, K.~M. 2023.
\newblock {Temporal Trends (from 2008 to 2017) in Functional Limitations and Limitations in Activities of Daily Living: Findings from a Nationally Representative Sample of 5.4 Million Older Americans.}
\newblock \emph{International journal of environmental research and public health}, 20(3).

\bibitem[{Grouvel et~al.(2023)Grouvel, Carcreff, Moissenet, and Armand}]{Grouvel2023}
Grouvel, G.; Carcreff, L.; Moissenet, F.; and Armand, S. 2023.
\newblock {A dataset of asymptomatic human gait and movements obtained from markers, IMUs, insoles and force plates}.
\newblock \emph{Scientific Data}, 10(1): 1--12.

\bibitem[{Halilaj et~al.(2018)Halilaj, Rajagopal, Fiterau, Hicks, Hastie, and Delp}]{Halilaj2018MachineOpportunities}
Halilaj, E.; Rajagopal, A.; Fiterau, M.; Hicks, J.~L.; Hastie, T.~J.; and Delp, S.~L. 2018.
\newblock {Machine learning in human movement biomechanics: Best practices, common pitfalls, and new opportunities}.
\newblock \emph{Journal of Biomechanics}, 81(16): 1--11.

\bibitem[{Heiden, Lloyd, and Ackland(2009)}]{Heiden2009}
Heiden, T.~L.; Lloyd, D.~G.; and Ackland, T.~R. 2009.
\newblock {Knee joint kinematics, kinetics and muscle co-contraction in knee osteoarthritis patient gait}.
\newblock \emph{Clinical Biomechanics}, 24(10): 833--841.

\bibitem[{Helseth, Hortob{\'{a}}gyi, and DeVita(2008)}]{Helseth2008}
Helseth, J.; Hortob{\'{a}}gyi, T.; and DeVita, P. 2008.
\newblock {How do low horizontal forces produce disproportionately high torques in human locomotion?}
\newblock \emph{Journal of Biomechanics}, 41(8): 1747--1753.

\bibitem[{Hulleck et~al.(2022)Hulleck, Menoth~Mohan, Abdallah, El~Rich, and Khalaf}]{Hulleck2022PresentTechnologies.}
Hulleck, A.~A.; Menoth~Mohan, D.; Abdallah, N.; El~Rich, M.; and Khalaf, K. 2022.
\newblock {Present and future of gait assessment in clinical practice: Towards the application of novel trends and technologies.}
\newblock \emph{Frontiers in medical technology}, 4: 901331.

\bibitem[{Hunter et~al.(2019)Hunter, Garcia, Shim, and Miller}]{Hunter2019FastRunning}
Hunter, J.~G.; Garcia, G.~L.; Shim, J.~K.; and Miller, R.~H. 2019.
\newblock {Fast Running Does Not Contribute More to Cumulative Load than Slow Running}.
\newblock \emph{Medicine and Science in Sports and Exercise}, 51(6): 1178--1185.

\bibitem[{Jacobs and Ferris(2015)}]{Jacobs2015EstimationSensors}
Jacobs, D.~A.; and Ferris, D.~P. 2015.
\newblock {Estimation of ground reaction forces and ankle moment with multiple , low-cost sensors}.
\newblock \emph{Journal of NeuroEngineering and Rehabilitation}, 1--12.

\bibitem[{Kingma and Ba(2014)}]{kingma2014adam}
Kingma, D.~P.; and Ba, J. 2014.
\newblock Adam: A method for stochastic optimization.
\newblock \emph{arXiv preprint arXiv:1412.6980}.

\bibitem[{Krupenevich et~al.(2015)Krupenevich, Rider, Domire, and Devita}]{Krupenevich2015MalesLoad}
Krupenevich, R.; Rider, P.; Domire, Z.; and Devita, P. 2015.
\newblock {Males and females respond similarly to walking with a standardized, heavy load}.
\newblock \emph{Military Medicine}, 180(9): 994--1000.

\bibitem[{Labban et~al.(2021)Labban, Stadnyk, Sommerfeldt, Nathanail, Dennett, Westover, Manaseer, and Beaupre}]{Labban2021KineticApproaches.}
Labban, W.; Stadnyk, M.; Sommerfeldt, M.; Nathanail, S.; Dennett, L.; Westover, L.; Manaseer, T.; and Beaupre, L. 2021.
\newblock {Kinetic measurement system use in individuals following anterior cruciate ligament reconstruction: a scoping review of methodological approaches.}
\newblock \emph{Journal of experimental orthopaedics}, 8(1): 81.

\bibitem[{Lein et~al.(2022)Lein, Alotaibi, Almutairi, and Singh}]{Lein2022Normative}
Lein, D.~H.; Alotaibi, M.; Almutairi, M.; and Singh, H. 2022.
\newblock Normative Reference Values and Validity for the 30-Second Chair-Stand Test in Healthy Young Adults.
\newblock \emph{International Journal of Sports Physical Therapy}, 17(5): 907--914.

\bibitem[{Lipton, Berkowitz, and Elkan(2015)}]{lipton2015criticalreviewrecurrentneural}
Lipton, Z.~C.; Berkowitz, J.; and Elkan, C. 2015.
\newblock A Critical Review of Recurrent Neural Networks for Sequence Learning.
\newblock arXiv:1506.00019.

\bibitem[{Lugaresi et~al.(2019)Lugaresi, Tang, Nash, McClanahan, Uboweja, Hays, Zhang, Chang, Yong, Lee, Chang, Hua, Georg, and Grundmann}]{Lugaresi2019MediaPipe:Pipelines}
Lugaresi, C.; Tang, J.; Nash, H.; McClanahan, C.; Uboweja, E.; Hays, M.; Zhang, F.; Chang, C.-L.; Yong, M.~G.; Lee, J.; Chang, W.-T.; Hua, W.; Georg, M.; and Grundmann, M. 2019.
\newblock {MediaPipe: A Framework for Building Perception Pipelines}.

\bibitem[{Lynn, Reid, and Costigan(2007)}]{Lynn2007TheAnalysis}
Lynn, S.~K.; Reid, S.~M.; and Costigan, P.~A. 2007.
\newblock {The influence of gait pattern on signs of knee osteoarthritis in older adults over a 5-11 year follow-up period: A case study analysis}.
\newblock \emph{Knee}, 14(1): 22--28.

\bibitem[{Maniar et~al.(2022)Maniar, Cole, Bryant, and Opar}]{Maniar2022MuscleLoading}
Maniar, N.; Cole, M.~H.; Bryant, A.~L.; and Opar, D.~A. 2022.
\newblock {Muscle Force Contributions to Anterior Cruciate Ligament Loading}.
\newblock \emph{Sports Medicine}, 52(8): 1737--1750.

\bibitem[{Martel-Pelletier et~al.(2016)Martel-Pelletier, Barr, Cicuttini, Conaghan, Cooper, Goldring, Goldring, Jones, Teichtahl, and Pelletier}]{Martel-Pelletier2016Osteoarthritis}
Martel-Pelletier, J.; Barr, A.~J.; Cicuttini, F.~M.; Conaghan, P.~G.; Cooper, C.; Goldring, M.~B.; Goldring, S.~R.; Jones, G.; Teichtahl, A.~J.; and Pelletier, J.~P. 2016.
\newblock {Osteoarthritis}.
\newblock \emph{Nature Reviews Disease Primers}, 2.

\bibitem[{Milner, Davis, and Hamill(2006)}]{Milner2006}
Milner, C.~E.; Davis, I.~S.; and Hamill, J. 2006.
\newblock Free moment as a predictor of tibial stress fracture in distance runners.
\newblock \emph{Journal of Biomechanics}, 39(15): 2819--2825.
\newblock Epub 2005 Nov 10.

\bibitem[{Miyazaki et~al.(2002)Miyazaki, Wada, Kawahara, Sato, Baba, and Shimada}]{Miyazaki2002DynamicOsteoarthritis}
Miyazaki, T.; Wada, M.; Kawahara, H.; Sato, M.; Baba, H.; and Shimada, S. 2002.
\newblock {Dynamic load at baseline can predict radiographic disease progression in medial compartment knee osteoarthritis}.
\newblock \emph{Annals of the Rheumatic Diseases}, 61(7): 617--622.

\bibitem[{Molavian et~al.(2023)Molavian, Fatahi, Abbasi, and Khezri}]{Molavian2023}
Molavian, R.; Fatahi, A.; Abbasi, H.; and Khezri, D. 2023.
\newblock Artificial Intelligence Approach in Biomechanics of Gait and Sport: A Systematic Literature Review.
\newblock \emph{Journal of Biomedical Physics and Engineering}, 13(5): 383--402.

\bibitem[{M{\"{u}}ndermann et~al.(2004)M{\"{u}}ndermann, Dyrby, Hurwitz, Sharma, and Andriacchi}]{Mundermann2004}
M{\"{u}}ndermann, A.; Dyrby, C.~O.; Hurwitz, D.~E.; Sharma, L.; and Andriacchi, T.~P. 2004.
\newblock {Potential Strategies to Reduce Medial Compartment Loading in Patients With Knee Osteoarthritis of Varying Severity: Reduced Walking Speed}.
\newblock \emph{Arthritis and Rheumatism}, 50(4): 1172--1178.

\bibitem[{Ngueleu et~al.(2019)Ngueleu, Blanchette, Bouyer, Maltais, McFadyen, Moffet, and Batcho}]{Ngueleu2019}
Ngueleu, A.~M.; Blanchette, A.~K.; Bouyer, L.; Maltais, D.; McFadyen, B.~J.; Moffet, H.; and Batcho, C.~S. 2019.
\newblock Design and Accuracy of an Instrumented Insole Using Pressure Sensors for Step Count.
\newblock \emph{Sensors}, 19(5): 984.

\bibitem[{{\"{O}}zate{\c{s}} et~al.(2024){\"{O}}zate{\c{s}}, Yaman, Salami, Campos, Wolf, and Schneider}]{Ozates2024IdentificationAI}
{\"{O}}zate{\c{s}}, M.~E.; Yaman, A.; Salami, F.; Campos, S.; Wolf, S.~I.; and Schneider, U. 2024.
\newblock {Identification and interpretation of gait analysis features and foot conditions by explainable AI}.
\newblock \emph{Scientific Reports}, 14(1): 5998.

\bibitem[{Ozdemir and Sonmez(2020)}]{Ozdemir2020WeightedImages}
Ozdemir, O.; and Sonmez, E.~B. 2020.
\newblock {Weighted Cross-Entropy for Unbalanced Data with Application on COVID X-ray images}.
\newblock \emph{Proceedings - 2020 Innovations in Intelligent Systems and Applications Conference, ASYU 2020}.

\bibitem[{Pattanapisont et~al.(2024)Pattanapisont, Kotani, Siritanawan, Kondo, and Karnjana}]{Pattanapisont2024Multi-ViewParts}
Pattanapisont, T.; Kotani, K.; Siritanawan, P.; Kondo, T.; and Karnjana, J. 2024.
\newblock {Multi-View Gait Analysis by Temporal Geometric Features of Human Body Parts}.
\newblock \emph{Journal of Imaging}, 10(4): 1--18.

\bibitem[{Rouhani et~al.(2010)Rouhani, Favre, Crevoisier, and Aminian}]{Rouhani2010AmbulatoryDistribution}
Rouhani, H.; Favre, J.; Crevoisier, X.; and Aminian, K. 2010.
\newblock {Ambulatory assessment of 3D ground reaction force using plantar pressure distribution}.
\newblock \emph{Gait and Posture}, 32(3): 311--316.

\bibitem[{Rynne et~al.(2022)Rynne, Le~Tong, Cheung, and Constantinou}]{Rynne2022EffectivenessMeta-analysis}
Rynne, R.; Le~Tong, G.; Cheung, R.~T.; and Constantinou, M. 2022.
\newblock {Effectiveness of gait retraining interventions in individuals with hip or knee osteoarthritis: A systematic review and meta-analysis}.
\newblock \emph{Gait and Posture}, 95(November 2021): 164--175.

\bibitem[{Savelberg and Lange(1999)}]{Savelberg1999a}
Savelberg, H.~H.; and Lange, A.~L. 1999.
\newblock {Assessment of the horizontal, fore-aft component of the ground reaction force from insole pressure patterns by using artificial neural networks}.
\newblock \emph{Clinical Biomechanics}, 14(8): 585--592.

\bibitem[{Sharma et~al.(1998)Sharma, Hurwitz, Thonar, Sum, Lenz, Dunlop, Schnitzer, Kirwan-Mellis, and Andriacchi}]{Sharma1998KneeOsteoarthritis}
Sharma, L.; Hurwitz, D.~E.; Thonar, E.~J.; Sum, J.~A.; Lenz, M.~E.; Dunlop, D.~D.; Schnitzer, T.~J.; Kirwan-Mellis, G.; and Andriacchi, T.~P. 1998.
\newblock {Knee adduction moment, serum hyaluronan level, and disease severity in medial tibiofemoral osteoarthritis}.
\newblock \emph{Arthritis and Rheumatism}, 41(7): 1233--1240.

\bibitem[{Sharma et~al.(2010)Sharma, Song, Dunlop, Felson, Lewis, Segal, Torner, Derek, Hietpas, Lynch, and Nevitt}]{Sharma2010}
Sharma, L.; Song, J.; Dunlop, D.; Felson, D.; Lewis, C.~E.; Segal, N.; Torner, J.; Derek, V.~C.; Hietpas, J.; Lynch, J.; and Nevitt, M. 2010.
\newblock {Varus and Valgus Alignment and Incident and Progressive Knee Osteoarthritis}.
\newblock \emph{Ann Rheum Dis.}, 69(11): 1940--1945.

\bibitem[{Sherstinsky(2020)}]{Sherstinsky_2020}
Sherstinsky, A. 2020.
\newblock Fundamentals of Recurrent Neural Network (RNN) and Long Short-Term Memory (LSTM) network.
\newblock \emph{Physica D: Nonlinear Phenomena}, 404: 132306.

\bibitem[{Shull et~al.(2013)Shull, Silder, Shultz, Dragoo, Besier, Delp, and Cutkosky}]{Shull2013Six-weekOsteoarthritis}
Shull, P.~B.; Silder, A.; Shultz, R.; Dragoo, J.~L.; Besier, T.~F.; Delp, S.~L.; and Cutkosky, M.~R. 2013.
\newblock {Six-week gait retraining program reduces knee adduction moment, reduces pain, and improves function for individuals with medial compartment knee osteoarthritis}.
\newblock \emph{Journal of Orthopaedic Research}, 31(7): 1020--1025.

\bibitem[{Sim et~al.(2015)Sim, Kwon, Oh, Joo, Choi, Heo, Kim, and Mun}]{Sim2015a}
Sim, T.; Kwon, H.; Oh, S.~E.; Joo, S.~B.; Choi, A.; Heo, H.~M.; Kim, K.; and Mun, J.~H. 2015.
\newblock {Predicting Complete Ground Reaction Forces and Moments During Gait With Insole Plantar Pressure Information Using a Wavelet Neural Network}.
\newblock \emph{Journal of Biomechanical Engineering}, 137(9): 1--9.

\bibitem[{Singh et~al.(2023)Singh, Bevilacqua, Nguyen, Hu, McGuinness, O’Reilly, Whelan, Caulfield, and Ifrim}]{Singh2023FastClassifiers}
Singh, A.; Bevilacqua, A.; Nguyen, T.~L.; Hu, F.; McGuinness, K.; O’Reilly, M.; Whelan, D.; Caulfield, B.; and Ifrim, G. 2023.
\newblock {Fast and robust video-based exercise classification via body pose tracking and scalable multivariate time series classifiers}.
\newblock \emph{Data Mining and Knowledge Discovery}, 37(2): 873--912.

\bibitem[{Snyder et~al.(2023)Snyder, Chu, Um, Heo, Miller, and Shim}]{Snyder2023a}
Snyder, S.~J.; Chu, E.; Um, J.; Heo, Y.~J.; Miller, R.~H.; and Shim, J.~K. 2023.
\newblock {Prediction of knee adduction moment using innovative instrumented insole and deep learning neural networks in healthy female individuals}.
\newblock \emph{Knee}, 41: 115--123.

\bibitem[{Spencer, Samaan, and Noehren(2023)}]{Spencer2023}
Spencer, A.; Samaan, M.; and Noehren, B. 2023.
\newblock {Monitoring Knee Contact Force with Force-Sensing Insoles}.
\newblock \emph{Sensors}, 23(10).

\bibitem[{Stenum et~al.(2024)Stenum, Hsu, Pantelyat, and Roemmich}]{Stenum2024ClinicalChange}
Stenum, J.; Hsu, M.~M.; Pantelyat, A.~Y.; and Roemmich, R.~T. 2024.
\newblock {Clinical gait analysis using video-based pose estimation: Multiple perspectives, clinical populations, and measuring change}.
\newblock \emph{PLOS Digital Health}, 3(3): e0000467.

\bibitem[{Thilakarathne et~al.(2022)Thilakarathne, Nibali, He et~al.}]{Thilakarathne2022}
Thilakarathne, H.; Nibali, A.; He, Z.; et~al. 2022.
\newblock Pose is all you need: the pose only group activity recognition system (POGARS).
\newblock \emph{Machine Vision and Applications}, 33: 95.

\bibitem[{Uhlrich et~al.(2023)Uhlrich, Falisse, Kidzinski, Muccini, Ko, Chaudhari, Hicks, and Delp}]{Uhlrich2023OpenCap:Videos}
Uhlrich, S.~D.; Falisse, A.; Kidzinski, L.; Muccini, J.; Ko, M.; Chaudhari, A.~S.; Hicks, J.~L.; and Delp, S.~L. 2023.
\newblock {OpenCap: Human movement dynamics from smartphone videos}.
\newblock \emph{PLoS Computational Biology}, 19(10 October): 1--26.

\bibitem[{van Lummel et~al.(2016)van Lummel, Walgaard, Maier, Ainsworth, Beek, and van Die{\"{e}}n}]{vanLummel2016TheAdults.}
van Lummel, R.~C.; Walgaard, S.; Maier, A.~B.; Ainsworth, E.; Beek, P.~J.; and van Die{\"{e}}n, J.~H. 2016.
\newblock {The Instrumented Sit-to-Stand Test (iSTS) Has Greater Clinical Relevance than the Manually Recorded Sit-to-Stand Test in Older Adults.}
\newblock \emph{PloS one}, 11(7): e0157968.

\bibitem[{Vaswani et~al.(2023)Vaswani, Shazeer, Parmar, Uszkoreit, Jones, Gomez, Kaiser, and Polosukhin}]{vaswani2023attentionneed}
Vaswani, A.; Shazeer, N.; Parmar, N.; Uszkoreit, J.; Jones, L.; Gomez, A.~N.; Kaiser, L.; and Polosukhin, I. 2023.
\newblock Attention Is All You Need.
\newblock arXiv:1706.03762.

\bibitem[{Winter(2009)}]{Winter2009}
Winter, D.~A. 2009.
\newblock \emph{{Biomechanics and Motor Control of Human Movement}}.
\newblock John Wiley {\&} Sons, Inc., fourth edi edition.
\newblock ISBN 9780470398180.

\bibitem[{Yoshiko et~al.(2021)Yoshiko, Ogawa, Shimizu, Radaelli, Neske, Maeda, Maeda, Teodoro, Tanaka, Pinto, and Akima}]{Yoshiko2021}
Yoshiko, A.; Ogawa, M.; Shimizu, K.; Radaelli, R.; Neske, R.; Maeda, H.; Maeda, K.; Teodoro, J.; Tanaka, N.; Pinto, R.~S.; and Akima, H. 2021.
\newblock Chair sit-to-stand performance is associated with diagnostic features of sarcopenia in older men and women.
\newblock \emph{Archives of Gerontology and Geriatrics}, 96: 104463.

\bibitem[{Yu and Garrett(2007)}]{Yu2007}
Yu, B.; and Garrett, W.~E. 2007.
\newblock Mechanisms of non-contact ACL injuries.
\newblock \emph{British Journal of Sports Medicine}, 41(Suppl 1): i47--i51.

\end{thebibliography}

\end{document}